# Semantic Segmentation–Driven Image-Level Diagnosis of Liver Cancers in Hematoxylin & Eosin Histopathology Images

Ivica Kopriva[a*], Dario Sitnik[b], Arijana Pačić[c], Karolina Krstanac[c], Irena Veliki Dalić[d], Ana Lukač[d], and Marijana Popović Hadžija[e]

[a]Division of Computing and Data Science, Ruđer Bošković Institute, Bijenička cesta 54, Zagreb Croatia

[b]Sitnik Ventures UG, Lautensackstr. 25, 80687 Munich, Germany

[c]Department of Pathology and Cytology, Dubrava University Hospital, Avenija Gojka Šuška 6, 10000 Zagreb, Croatia

[d]Clinical Division for Pathology and Cytology, Clinical Hospital Center Zagreb, Kišpatićeva 12, 10000 Zagreb, Croatia

[e]Division of Molecular Medicine, Ruđer Bošković Institute, Bijenička cesta 54, 10000 Zagreb, Croatia

[*]Corresponding author

*E-mail addresses*: ikopriva@irb.hr (I.Kopriva), dario.sitnik@gmail.com (D. Sitnik), arijanapacic@yahoo.com (A. Pačić), karolina0101@windowslive.com (K. Krstanac), irenavelikidalic@gmail.com (I. Veliki Dalić), lukac_ana@hotmail.com (A. Lukač), Marijana.Popovic.Hadzija@irb.hr (M. Popović Hadžija).

## Abstract

Computational analysis of histopathological images plays a central role in cancer diagnosis and treatment planning. As hematoxylin and eosin (H&E) staining constitutes the primary entry point in routine diagnostic workflows, computer-aided diagnosis from whole-slide H&E images is of particular clinical relevance. However, substantial variability in specimen preparation, staining protocols, and scanning conditions, together with inherent uncertainty in expert pixel-level annotations, makes automated analysis of H&E-stained images challenging. In this study, we propose a semantic segmentation–based framework for image-level diagnosis, grounded in the clinically motivated assumption that each histopathological image corresponds to a single cancer type. Image-level predictions are obtained by assigning the class of the dominant pixel-level label in the segmentation output. To ensure clinical relevance, we adopt the nnU-Net architecture and train it on a publicly available dataset collected in our study with pixel-level annotations for three liver cancer types: hepatocellular carcinoma (HCC; 55 images from 30 patients), cholangiocellular carcinoma (CCA; 55 images from 29 patients), and colorectal metastatic adenocarcinoma (CMA; 60 images from 30 patients). Annotations were independently provided by four pathologists, and consensus ground truth was established via majority voting with at least two out of four pathologists had to agree on cancer label. Inter-annotator agreement, measured using Fleiss' kappa, was 0.694 for HCC, 0.728 for CCA, and 0.756 for CMA, indicating substantial agreement. We hypothesize that the combination of stain normalization and semantic segmentation mitigates domain shift and reduces sensitivity to annotation noise, thereby limiting the need for model fine-tuning when new data are introduced. Five-fold cross-validation yielded balanced accuracies of 0.975 (HCC), 0.950 (CCA), and 1.000 (CMA), comparable to results obtained with immunohistochemical staining and superior to several deep learning models trained on patch-level annotations. On intraoperatively acquired CMA samples not included in the training cohort, nnU-Net achieved a balanced accuracy of 0.947, demonstrating robustness to domain shift that was not case with models trained on patch-level annotations. The proposed framework has the potential to support pathologists in prioritizing immunohistochemical marker selection, thereby reducing diagnostic costs and turnaround time. Integration with immunohistochemical findings may further improve overall diagnostic reliability. The code with explanations is available at: `https://github.com/dsitnik/liver3c`. Data are available at: `https://data.fulir.irb.hr/data/HandE-Liver3C/HandE-Liver3C_data.zip`. Weights of trained networks are available at: `https://data.fulir.irb.hr/data/HandE-Liver3C/HandE-Liver3C_weights.zip`.

*Keywords*: Computational pathology, H&E histopathology, Liver cancer, Semantic segmentation, Image-level diagnosis.

## 1.0 Introduction

Pathology and computational pathology image analysis play a pivotal role in the diagnosis and treatment of cancer (Yu et al., 2018; Zhang et al., 2025; Song et al., 2023; Xing and Jang, 2016). Among histopathological modalities, hematoxylin and eosin (H&E)-stained specimens remain the most widely used in routine clinical practice (Fischer et al., 2008). In H&E staining, hematoxylin colors acidic cellular components, particularly nuclei, in blue-purple hues, whereas eosin stains basic structures such as cytoplasm and extracellular matrix in varying shades of pink. Primary liver cancer ranks as the sixth most frequently diagnosed cancer worldwide and the fourth leading cause of cancer-related mortality (Hu et al., 2025). The predominant histological subtypes are hepatocellular carcinoma (HCC) and intrahepatic cholangiocellular carcinoma (CCA), accounting for approximately 75–85% and 10–15% of cases, respectively (Sung et al., 2021; Hu et al., 2025). In addition to primary tumors, the liver is a common site of metastasis, particularly from colorectal cancer. According to a recent report from the National Cancer Institute (2025), colorectal cancer accounts for 8.6% of all cancer-related deaths in the United States. A large SEER-based analysis including 1,630,725 cancer cases identified 105,329 patients with liver metastases at diagnosis, of which 16% originated from colorectal adenocarcinoma (Wang et al., 2020). Consequently, colorectal metastatic adenocarcinoma (CMA) represents a clinically relevant differential diagnosis in liver pathology. Despite the high clinical burden, only a limited number of studies have addressed automated differentiation between HCC and CCA in H&E-stained slides. Hägele et al. (2024) combined semantic segmentation with patch-based weak complementary labels during training, while Kiani et al. (2020) employed patch-level annotations for binary discrimination between HCC and CCA. Most existing machine learning studies instead focus on distinguishing benign from malignant HCC tissue within liver (Chen et al., 2020). To the best of our knowledge, no prior work has investigated automated differentiation among HCC, CCA, and CMA using H&E-stained histopathological images.

Generally, H&E stains the nucleus and cytoplasm in both healthy and tumor cells. However, characteristic morphological microanatomy patterns, visible after staining, frequently guide pathologists in distinguishing normal from cancer tissue (Chan and Yeh, 2010). This observation motivates the development of automated analysis methods for H&E-stained histopathological images (Zhang et al., 2025), as part of broader efforts in medical image segmentation. Image segmentation constitutes a central component of computer-aided diagnosis systems and supports numerous clinical applications, including diagnostic decision support (Fauw et al., 2018), intraoperative assistance (Hollon et al., 2020), tumor progression monitoring (Kickingereder et al., 2019), and image-guided interventions (Tonutti et al., 2017).

Image segmentation methods are commonly categorized into semantic and instance segmentation (Taghanaki et al., 2021; Minaee et al., 2022). Semantic segmentation performs pixel-level classification by assigning each pixel to a predefined category, thereby simplifying image representation and enabling structured tissue characterization (Rijthoven et al., 2021). In medical imaging, such spatially organized label maps facilitate downstream diagnostic analysis (Azad et al., 2024). Fully supervised semantic segmentation models have demonstrated strong performance (Graham et al., 2019); however, they require dense pixel-level annotations, which are time-consuming and labor-intensive to obtain in histopathology (Kopriva et al., 2025). Moreover, supervised models are susceptible to domain shift, often exhibiting degraded performance on data acquired under different staining, scanning, or preparation conditions (Marini et al., 2021). Consequently, achieving robust generalization typically requires large and diverse annotated datasets. Despite these challenges, fully supervised segmentation approaches have achieved high accuracy in normal and pathological tissue structure delineation, which is directly relevant to higher-level clinical tasks such as cancer diagnosis (Lou et al., 2023). Building on this evidence, we propose a semantic segmentation–based framework for image-level diagnosis of liver cancers in H&E-stained specimens. Under the clinically motivated assumption that each specimen corresponds to a single tumor type, image-level predictions are derived by aggregating pixel-level classifications and assigning the diagnosis according to the most prevalent class (HCC, CCA, or CMA). Although supervised segmentation methods are generally sensitive to domain shift, we observe that combining stain normalization (Reinhard et al., 2001) with pixel-level aggregation yields robust diagnostic performance, approaching that reported for immunohistochemical evaluation and exceeding the performance of five deep learning models trained on patch-level annotations. Given the known difficulty of differentiating primary liver cancers such as HCC and CCA—even with immunohistochemistry (Chan and Yeh, 2010)—these findings underscore the potential clinical utility of the proposed approach.

In histopathology, annotation quality is inherently dependent on the individual annotator, and achieving consistent labeling across experts remains challenging. Previous studies have reported that even among experienced pathologists, complete agreement in independent diagnoses may not exceed 75% (Elmore, 2015). Moreover, uncertainty in diagnostic labeling can arise even at the slide level, particularly when regions exhibit features corresponding to multiple histological classes (Mercan et al., 2018). Such variability limits the reliability of supervised learning approaches and increases the annotation burden placed on pathologists (Yang et al., 2022). To address inter-observer variability, in our study four experienced pathologists independently performed pixel-level annotations of H&E-stained histopathological images corresponding to three types of liver cancer. Final ground truth labels were established using majority voting with at least two out of four pathologists had to agree on cancer label. Inter-annotator agreement was quantified using Fleiss' kappa (Fleiss, 1971; Landis and Koch, 1977) on approximately 60 images per class (CMA, HCC, and CCA), yielding values of 0.756, 0.694, and 0.728, respectively. According to established interpretation guidelines, these values indicate substantial agreement and are consistent with previously reported levels of diagnostic concordance (Elmore, 2015). In addition to the domain shift associated with data heterogeneity, annotation variability represents a further source of uncertainty in supervised training. We hypothesize that aggregating pixel-level predictions into image-level diagnoses may reduce sensitivity to local annotation inconsistencies, thereby improving robustness to unavoidable imperfections in expert-provided labels.

The task-specific design and configuration of deep learning–based segmentation models often require substantial methodological expertise, which can limit their broader applicability across diverse image analysis problems (Litjens et al., 2017). To address this challenge, Isensee et al. (2021) introduced nnU-Net, an open-source framework that automatically adapts preprocessing, network architecture, training, and post-processing pipelines to a given biomedical segmentation task. Building upon the widely adopted U-Net architecture (Azad et al., 2024), nnU-Net systematizes empirical best practices in model configuration and provides a standardized baseline for new segmentation problems. Beyond serving as a competitive reference method, nnU-Net facilitates reproducibility and comparability across datasets and applications. Its adaptability has supported evaluation on a wide range of biomedical segmentation tasks, including recent applications in organ and plaque segmentation from CT imaging (Liu et al., 2026). Motivated by these properties, we employ nnU-Net as the backbone of our semantic segmentation framework for image-level diagnosis. To mitigate variability in staining appearance and batch effects inherent to H&E histopathology, we additionally apply the stain normalization method proposed by Reinhard et al. (2001). The combination of automated configuration and standardized preprocessing enhances methodological transparency and supports reproducible deployment. Together with the publicly available annotated dataset, this framework provides a basis for systematic benchmarking and further development of automated liver cancer diagnosis from H&E-stained images. Proposed methodology is illustrated in Fig. 1.

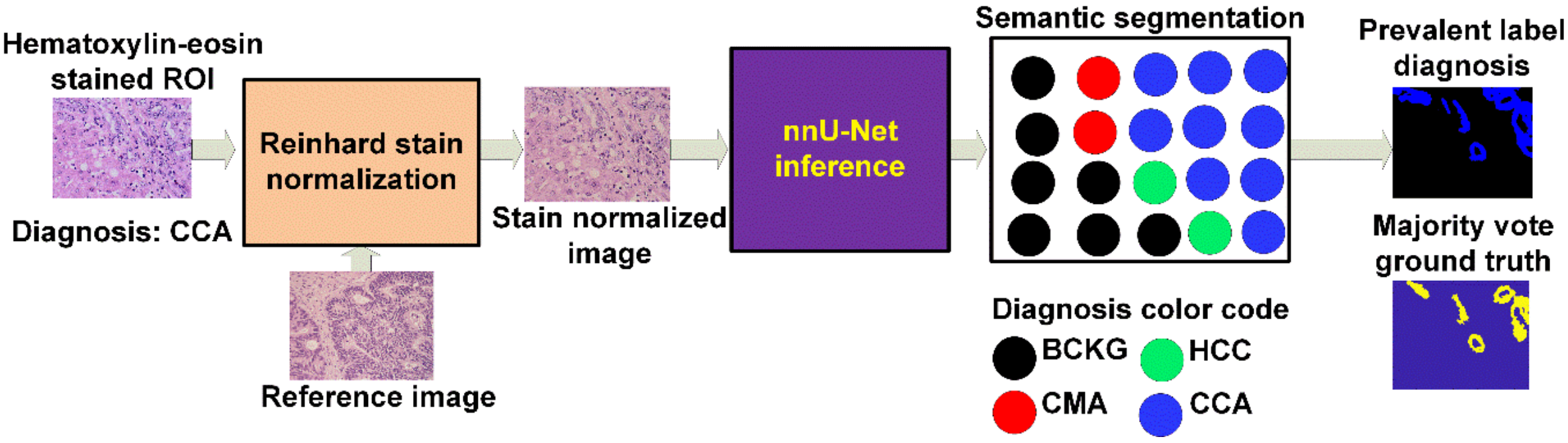

**Fig. 1**. Overview of the proposed semantic segmentation–based framework for image-level diagnosis of liver cancers. H&E-stained regions of interest are first subjected to stain normalization to reduce appearance variability. A semantic segmentation model predicts pixel-level labels corresponding to colorectal metastatic adenocarcinoma (CMA), hepatocellular carcinoma (HCC), cholangiocellular carcinoma (CCA), and background tissue (BCKG). Image-level diagnosis is subsequently derived by aggregating pixel-level predictions and assigning the class with the highest prevalence within the region of interest.

Motivated by the considerations outlined above, the main contributions of this work are as follows:

- **Publicly available pixel-level annotated dataset of liver cancers**. We present a dataset comprising 170 images of H&E-stained histopathological samples from three liver cancer categories: hepatocellular carcinoma (HCC; 55 images from 30 patients), cholangiocellular carcinoma (CCA; 55 images from 29 patients), and colorectal metastatic adenocarcinoma (CMA; 60 images from 30 patients). Pixel-level annotations were independently provided by four experienced pathologists, and consensus ground truth was established via majority voting with at least two out of four pathologists had to agree on cancer label. To the best of our knowledge, this is the only publicly available dataset containing pixel-wise annotations for these three liver cancer types. In addition to that, we present an external clinical test dataset, not used for network training, comprising images of 102 H&E-stained histopathological samples from three liver cancer categories: HCC (22 images from 11 patients), CCA (21 image from 11 patients), and CMA (40 images from 20 patients from formalin-fixed paraffin-embedded sections, and 19 images of samples collected intraoperatively from 19 patients). The datasets are accessible at: `https://data.fulir.irb.hr/data/HandE-Liver3C/HandE-Liver3C_data.zip`.

- **Semantic segmentation–driven image-level diagnosis**. We propose a framework that derives image-level diagnoses by aggregating pixel-level semantic segmentation outputs. Under the clinically motivated assumption that each specimen corresponds to a single tumor type, the final diagnosis is assigned according to the most prevalent predicted class. This aggregation strategy aims to reduce sensitivity to local annotation inconsistencies and domain variability. The approach is directly applicable to regions of interest extracted from whole-slide images.
- **Interpretable tumor localization**. In addition to image-level classification, the proposed method provides spatially resolved tumor localization via semantic segmentation maps. This supports clinical interpretability and enables visual assessment of diagnostically relevant regions.
- **Standardized and reproducible implementation**. The framework is implemented using the nnU-Net architecture combined with stain normalization following Reinhard et al. (2001) to mitigate appearance variability in H&E images. This standardized configuration enhances reproducibility and facilitates deployment without extensive task-specific re-engineering. The trained inference model is publicly available at: `https://data.fulir.irb.hr/data/HandE-Liver3C/HandE-Liver3C_weights.zip`, and the source code is publicly available at: `https://github.com/dsitnik/liver3c`.
- **Clinical benchmarking against immunohistochemical evaluation and patch-level annotated images**. We demonstrate that the proposed approach achieves diagnostic performance comparable to that reported for immunohistochemical staining and exceeding that of five deep learning models trained on patch-level annotations. These findings suggest that the method may assist pathologists in prioritizing immunohistochemical marker selection and could complement established diagnostic workflows.
- **Clinical benchmarking against patch-level annotated images on intraoperatively acquired CMA samples**. On intraoperatively acquired CMA samples not included in the training cohort, nnU-Net achieved a balanced accuracy of 0.947 (see confusion matrix in Table 8), demonstrating robustness to domain shift that was not case with patch-level annotated models.

## 2. Related work

### *2.1. Liver cancer datasets with pixel-level annotations*

A few large, locally annotated datasets have been developed in computational pathology and have been widely used in fully supervised approaches for prostate cancer (Arvaniti et al., 2018; Ström et al., 2019; Nagpal et al., 2019). To date, only one publicly available pixel-level annotated dataset exists for liver cancer (Kim et al., 2021). The PAIP liver cancer dataset comprises 50 whole-slide images of HCC resection specimens with pixel-level annotations for total tumor mass, viable tumor, and necrosis regions, along with associated clinical outcomes. Collected at Seoul National University Hospital, the dataset consists of high-resolution H&E-stained slides (0.25 μm/pixel) scanned at 20× magnification and annotated by two pathologists. Additional pixel-level annotated datasets of H&E-stained primary liver tumor resections, including HCC and CCA, have been reported by Hägele et al. (2024), but these are derived from the tissue bank of Charité Universitätsmedizin Berlin and are not publicly accessible. In contrast, our dataset represents the only dataset to date providing pixel-level annotations for H&E-stained images covering three liver cancer types: two types of primary liver cancers (HCC and CCA), and metastatic colorectal carcinoma in the liver (CMA).

### *2.2. Source variability and domain shift*

An important practical challenge in computational histopathology is source variability, arising from differences in staining and imaging protocols, tissue section thickness, patient cohorts, and other factors (Kumari et al., 2025; Hetz et al., 2024). Such variability contributes to heterogeneity in histopathological data and poses a critical challenge for the generalizability of deep learning models, particularly given the limited availability of large, diverse, and fully annotated datasets for training (Zhou et al., 2021; Marini et al., 2021; Guan and Liu, 2022). Differences between training and test datasets can result in notable performance degradation (Zhou et al., 2021). While domain adaptation methods have been proposed to address generalization issues, they are often sensitive to the complex characteristics of healthcare data (Azad et al., 2024). For instance, access to source data may be temporally limited due to privacy regulations, restricting the applicability of methods that require simultaneous access to source and target datasets (Thandiackal et al., 2024). Few-shot learning approaches have been explored to mitigate annotation burden in nucleus segmentation, leveraging the growing availability of fully annotated public datasets to transfer knowledge to target datasets with limited labels (Ming et al., 2025). In the present study, we address source variability by applying the stain normalization method of Reinhard et al. (2001), which standardizes H&E-stained images from multiple centers relative to a selected target image. The target image together with stain normalization code in MATLAB script language is available at: `https://github.com/dsitnik/liver3c/tree/main/scripts/preprocessing/stain_normalization`. Although simple, this preprocessing step enabled nnU-Net (Isensee et al., 2021) to achieve competitive diagnostic performance for liver cancer image-level classification (see section 4.4 for details). In addition to that we conjecture that aggregation from pixel-level classification to image-level diagnosis also improves robustness to domain shift problem (see section 4.6 for details related to intraoperatively collected CMA samples).

### *2.3. Immunohistochemical staining*

Ma et al. (1993) reported that distinguishing primary hepatocellular carcinoma (HCC) from metastatic carcinomas involving the liver is particularly challenging. Histochemical and immunohistochemical staining thus play a central role in the histopathological workflow (Rivenson et al., 2019). Early studies investigating specific IHC profiles to differentiate HCC, cholangiocellular carcinoma (CCA), and metastatic adenocarcinoma (MA) used panels of seven markers, achieving maximum positivity rates of approximately 71%. In these studies, CCAs frequently exhibited staining patterns similar to those of metastatic carcinomas (Lei et al., 2006). Marker-specific investigations have provided further insights. Lei et al. (2006) analyzed liver resections using TTF-1, CK19, HepPar-1, alpha-fetoprotein, polyclonal carcinoembryonic antigen, CK7, and CK20. TTF-1 cytoplasmic staining combined with CK19 emerged as potentially useful for distinguishing primary HCC from CCA and MA, although no marker alone achieved 100% sensitivity or specificity. Similarly, Balaton et al. (1988) reported universal CK19 negativity in HCC and universal positivity in CCA and MA, whereas Geramizadeh et al. (2007) found HepPar-1 positivity in 30/35 HCCs and absence in CCAs and MAs; MOC31 was positive in 60/65 MAs but negative in 30/45 HCCs. Hanif et al. (2014) corroborated HepPar-1 findings, and Stroescu et al. (2006) emphasized that expression patterns of CKsand carcinoembryonic antigen could aid differential diagnosis in challenging cases. Maeda et al. (1996) conducted IHC on 90 resected liver tumors (30 HCC, 30 CCA, 30 CMA) using CK7, CK19, and CK20. CK7 was diffusely positive in 97% of CCAs, 7% of HCCs, and 3% of CMAs; CK19 in 77% of CCAs, 64% of CMAs, and 0% of HCCs; CK20 in 74% of CMAs, 10% of CCAs, and 0% of HCCs. These cohort sizes match those used in the present study. Additional studies have evaluated CD10 (Ahuja et al., 2008), BSEP and MDR3 (Fujikura et al., 2016), and multiple markers including HepPar-1, CK7, and CK20 (Lau et al., 2002), consistently reporting partial overlap between tumor types but no single marker with perfect specificity. Chan and Yeh (2010) provide a comprehensive summary of nine IHC markers for differentiating HCC, CCA, and MA, confirming that no marker achieves 100% positivity for any single tumor type (see Table 7). The results obtained by the proposed nnU-Net–based semantic segmentation approach for image-level diagnosis are consistent with these immunohistochemical findings, demonstrating similar patterns of differentiation among HCC, CCA, and CMA and the absence of any marker specific to a single tumor type (see Table 2).

### *2.4. Automatic analysis of images of H&E-stained specimens*

In recent years, numerous approaches have been proposed for the automatic analysis of images of H&E-stained histopathological specimens. High model accuracy typically requires a substantial number of training labels, which incurs considerable time and effort from expert pathologists. To address this challenge, several annotation-efficient strategies have been developed. Multiple-instance learning (MIL) has become widely adopted in computational pathology because it requires only slide-level annotations (Laleh et al., 2022; Li et al., 2023). However, MIL methods generally rely on a large number of slides and focus on the most discriminative patches, achieving good slide-level accuracy but limited patch-level performance (Zhang et al., 2025). Improved patch-level classification models have been proposed to mitigate this limitation (Zhang et al., 2025; Geijs et al., 2024; Han et al., 2022). Few-shot learning approaches aim to reduce annotation requirements by leveraging the growing availability of fully annotated public datasets, which can be combined with target datasets containing limited labels (Ming et al., 2025). Active learning paradigms similarly reduce labeling costs by selecting the most informative samples from known classes for annotation (Tang et al., 2025). Vision-language models have also been applied to transfer nuclei-related prompts into representation learning for nuclei detection tasks, addressing morphological variability (Yao et al., 2025). Graph neural networks (GNNs) have been applied in histopathology to model relationships between cells, patches, or tissue regions, with nodes representing cells, patches, or higher-level semantic entities (Li et al., 2018; Brussee et al., 2025). These methods, however, are computationally intensive for pixel-level inputs due to the need for prior segmentation. Semi-supervised learning leverages limited labeled data alongside abundant unlabeled data to improve performance, as demonstrated in nucleus segmentation (Ke et al., 2023). Weakly supervised approaches utilize less expensive annotations, such as image-level labels, though they still require a large number of slides for effective training (Zhou et al., 2023). Self-supervised learning, a subclass of unsupervised methods, pre-trains models using the inherent structure of the data itself, enabling generalization to downstream tasks with limited annotations (Wang et al., 2021; Ciga et al., 2022). This approach has emerged as a promising solution to reduce annotation requirements while maintaining robust performance across diverse histopathological applications. Another annotation efficient method was proposed in Xu et al. (2026). It introduces a vector quantization-based pseudo-label generation mechanism and a cross-latent graph neural network for context aware relation modeling for scribble- and point-based weak annotations in clinical scenarios.

## 3. Material and methods

### *3.1. Database with pixel-level annotated images of liver cancers*

The liver cancer dataset comprises H&E-stained histopathological images from three cancer types: colorectal metastatic adenocarcinoma (CMA; 60 images from 30 patients), hepatocellular carcinoma (HCC; 55 images from 30 patients), and cholangiocellular carcinoma (CCA; 55 images from 29 patients). Initially, 60 images were collected for each group. Following pixel-level annotation by four pathologists, inter-annotator agreement was evaluated using Fleiss' kappa statistics for each image. Images with low agreement—kappa < 0.3 for HCC and < 0.4 for CCA—were excluded from the main dataset. Due to their challenging nature, a subset of these excluded images was retained for illustrative purposes to demonstrate diagnostic and localization performance of the trained nnU-Net model (see section 4.7). Approximately half of the images contained cancerous regions spanning the entire field of view, while in the remaining images, malignant regions occupied only a portion of the field. Samples were collected retrospectively between February 2024 and March 2026 at the Department of Pathology and Cytology, Dubrava University Hospital (DUH), and at the Clinical Division for Pathology and Cytology, Clinical Hospital Center (CHC), both in Zagreb, Croatia. Ethical approval for sample collection was obtained from the Institutional Review Board of CHC on September 22, 2022 (Class: 8.1.-22/177-2; Number 02/013/JG) and from DUH on August 29, 2022 (Approval No.: 2002/2908-02). Because of the retrospective nature of the study, informed consent was not required; all data were anonymized. Pathologist established diagnosis of primary hepatocellular carcinoma, cholangiocarcinoma and metastatic colonic adenocarcinoma in a liver after microscopic examination of samples stained by H&E and immunohistochemically stained to diagnosis-relevant antigens using primary antibodies specific to: hepatocyte protein (Hep Par 1 – OCH1E5) marker for HCC, cytokeratin 7 (CK7 - clone OV-TL 12/30) and cytokeratin 20 (CK20, clone Ks20.8) as markers of cholangiocaricnoma and adenocarcinoma cells, and Caudal related homebox transcription factor 2 (CDX2) expressed in colorectal carcinoma cells, all from Dako, Denmark.

### *3.1.1. Image acquisition*

As highlighted by Riasatian et al. (2021), histopathological images should be acquired at high magnification (≥20×) and with sufficiently large patches (e.g., 500 × 500 μm², corresponding to approximately 1000 × 1000 pixels at a pixel size of 0.25 μm²) to capture the features relevant for most diagnostic cases. In our dataset, images were acquired at 40× magnification and afterwards down sampled to 20× magnification with dimensions of 884 × 1124 pixels for CCA, 883 × 1224 pixels for HCC, and 1015 × 1920 pixels for CMA. The HCC and CCA images were acquired on the Olympus SC180 microscope with a final spatial resolution (after down sampling) of approximately 0.23 μm per pixel. The CMA images were acquired on the Zeiss Axiocam208 microscope with a final spatial resolution (after down sampling) of approximately 0.17 μm per pixel. Therefore, the acquired images satisfy recommended specifications.

### *3.1.2. Image annotation*

As highlighted in Section 1, annotation quality is highly dependent on the individual pathologist, and maintaining consistency across experts is challenging. Elmore (2015) reported that even experienced pathologists achieved only 75% unanimous agreement on independent diagnoses. To address potential inconsistencies in our study, four pathologists (A. P., I. V. D., A. L., and K. K.) performed pixel-level annotations of images of H&E-stained specimens corresponding to three types of liver cancers. Annotation was conducted using a superpixel-based software system, as described in Sitnik et al. (2021). Following individual annotations, the final ground truth was determined by majority voting with at least two out of four pathologists had to agree on cancer label. Annotators could label images either with a freehand brush tool or by assigning labels to superpixels produced by several alternative grouping algorithms, and were instructed not to discuss individual cases with one another so that the four annotations remained independent. Inter-annotator agreement was quantified using Fleiss' kappa statistics (Fleiss, 1971; Landis and Koch, 1977) for each cancer group separately. Table 1 presents intra-annotator kappa statistics reproduced from Fleiss (1971). To mitigate the influence of unreliably annotated images on model training, images with kappa ≤ 0.3 for HCC and ≤ 0.4 for CCA were excluded. Figure 2 illustrates the kappa statistics per image for each cancer type. The resulting inter-annotator agreement for CMA, HCC, and CCA was 0.756, 0.694, and 0.728, respectively, indicating substantial agreement according to Landis and Koch (1977). These values are consistent with the findings reported by Elmore (2015). Therefore, in addition to the previously discussed domain shift caused by data heterogeneity, we expect that aggregating pixel-level predictions from semantic segmentation to image-level diagnosis reduces the impact of unavoidable imperfections in expert-based annotations.

**Table 1.** Kappa statistics and strength of agreement in Fleiss (1971).

| Kappa statistics | Strength of Agreement |
|---|---|
| <0.00 | Poor |
| 0.00-0.20 | Slight |
| 0.21-0.40 | Fair |
| 0.41-0.60 | Moderate |
| 0.61-0.80 | Substantial |
| 0.81-1.00 | Almost Perfect |

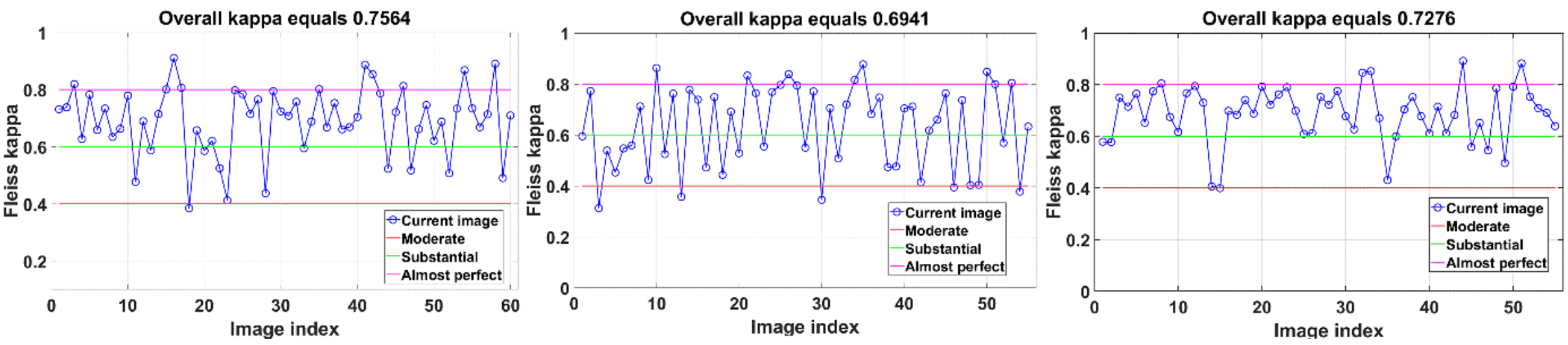

**Fig. 2**. The Fleiss' kappa statistics estimated for each image for pixel-wise annotations by four pathologists. From left to right: CMA, HCC, CCA.

### *3.2. Methodology*

This section introduces our proposed nnU-Net based semantic segmentation approach to image-level diagnosis as well as patch-level annotated approach to image-level diagnosis.

*3.2.1. nnU-Net semantic segmentation approach to image-level diagnosis*

As illustrated in Fig. 1, the selected region of interest (ROI) is first subjected to stain normalization using the method of Reinhard et al. (2001), which performs color standardization with respect to a predefined reference image. Although simple, this preprocessing step reduces substantially the heterogeneity of images acquired from the two hospital centers. The stain-normalized ROI is then processed by a trained nnU-Net model for semantic segmentation (pixel-level classification). We selected nnU-Net because it is an automated deep-learning framework specifically designed for biomedical image segmentation. The framework automatically derives all pipeline decisions: architecture topology, patch size, batch size, normalization strategy, and post-processing—all from a dataset fingerprint computed before training, and has been widely adopted as a strong baseline for segmentation tasks. Given 2D image inputs, it selects a 2D U-Net configuration with encoder-decoder structure, skip connections, instance normalization, and a fixed training recipe (Dice coefficient plus cross-entropy loss, SGD with Nesterov momentum 0.99, polynomial learning rate decay, and comprehensive augmentation); all weights are initialized from scratch, with no pretrained weights supplied. A four-class linear head (for the background, CMA, HCC, and CHO class) replaces the original ImageNet classifier. At inference, 384×384 pixel-patches are tiled non-overlappingly across each slide, and the image diagnosis is determined by majority vote, i.e., the most frequent tumor-class prediction across all patches. The nnU-Net, operates as a classification-via-segmentation model. It predicts a dense per-pixel class map and the image diagnosis is determined by majority vote among its positive (non-background) pixel predictions; at inference it applies sliding-window prediction with 50% overlap. nnU-Net training and inference use the nnU-Netv2 command-line interface (Isensee et al., 2021).

Despite stain normalization, residual variability in H&E-stained images and unavoidable imperfections in expert-based pixel-level annotations may still introduce errors during inference. To mitigate these effects, we exploited a clinically motivated prior assumption that each histopathological specimen corresponds to a single cancer type. Consequently, the final diagnosis is obtained by aggregating pixel-level predictions into an image-level decision using the concept of the dominant label. Specifically, an ROI is classified as healthy background if no pixels are predicted as cancerous. Otherwise, the ROI is assigned the cancer type corresponding to the class with the largest number of labeled pixels. This aggregation from pixel-level segmentation to image-level diagnosis increases robustness to local misclassifications caused by staining variability or annotation noise. Examples of color-coded semantic segmentation outputs produced by the trained nnU-Net inference model for the diagnosis of CMA, HCC, and CCA are shown in Fig. 3. In addition to providing an image-level diagnosis, the resulting segmentation maps enable tumor localization, thereby improving the interpretability of the model predictions.

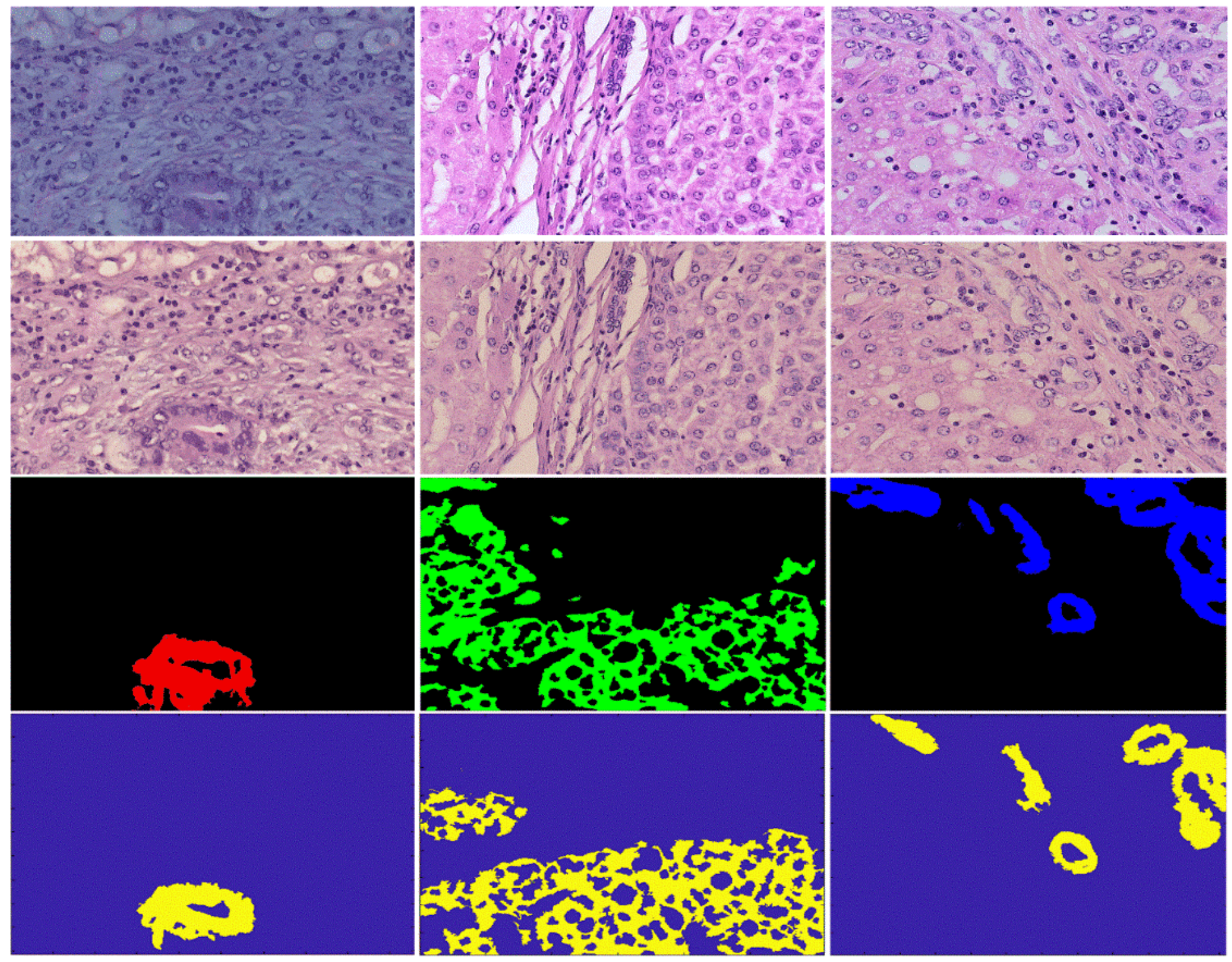

**Fig. 3**. Illustration of the diagnostic pipeline using the nnU-Net inference model. Columns correspond to different cancer types (from left to right): colorectal metastatic adenocarcinoma (CMA), hepatocellular carcinoma (HCC), and cholangiocellular carcinoma (CCA). Rows illustrate successive processing steps: (1) original H&E-stained images; (2) stain-normalized images obtained using the Reinhard method; (3) semantic segmentation outputs produced by the nnU-Net inference model; and (4) ground-truth annotations obtained by majority voting from pixel-level annotations provided by four pathologists with at least two out of four pathologists had to agree on cancer label. Color coding of predicted classes: red—CMA, green—HCC, blue—CCA.

### *3.2.2. Patch-level labeling for image-level diagnosis*

As discussed in Section 2.4, acquiring a large number of training labels in computational pathology requires substantial time and effort from human experts, particularly when pixel-level annotations are required. Several strategies have therefore been proposed to reduce annotation costs. For example, multiple instance learning (MIL) requires only slide-level annotations (Laleh et al., 2022; Li et al., 2023), while patch-level classification approaches rely on annotations at the level of image patches rather than individual pixels (Zhang et al., 2025; Geijs et al., 2024; Han et al., 2022). In the context of liver cancer classification, Hägele et al. (2024) combined semantic segmentation with patch-based weak complementary labels during training, whereas Kiani et al. (2020) used patch-level annotations to discriminate between HCC and CCA. Furthermore, the uncertainty observed in pixel-level annotations by pathologists provides additional motivation for patch-level annotation, where labeling inconsistencies may be reduced. To provide a comparative baseline for our proposed semantic segmentation framework, we trained and evaluated five deep neural networks using patch-level annotated images: ConvNeXtV2 (Woo et al., 2023), DenseNet (Huang et al., 2017), EfficientNetV2 (Tan and Le, 2021), MaxViT (Tu et al., 2022), and SwinV2 (Liu et al., 2021; Liu et al., 2022). All five models have a hierarchical encoder backbone and are fine-tuned from ImageNet pre-trained weights with a common approach: AdamW (Loshchilov and Hutter, 2019) with weight decay 0.01, initial learning rate $10^{-4}$ halved after 10 epochs without improvement on validation Dice (minimum $10^{-7}$), weighted cross-entropy loss with inverse-frequency class weights, gradient clipping at $\ell_2$ norm 1.0, and early stopping after 20 epochs without improvement over a maximum of 500 epochs. During training, patches for these five models are extracted with 50% overlap to increase sample density. Patches whose tumor-positive fraction falls strictly between 0% and 10% are discarded to avoid ambiguous boundary examples. Class imbalance is addressed at two levels: background patches are capped at twice the mean cancer-class count per case, and a weighted random sampler applies inverse-frequency class weights at the batch level during training. Patch-level predictions were subsequently aggregated to obtain image-level diagnoses for each ROI. Fig. 4 illustrates representative patch-level diagnostic maps for three ROI images diagnosed as CMA, HCC, and CCA.

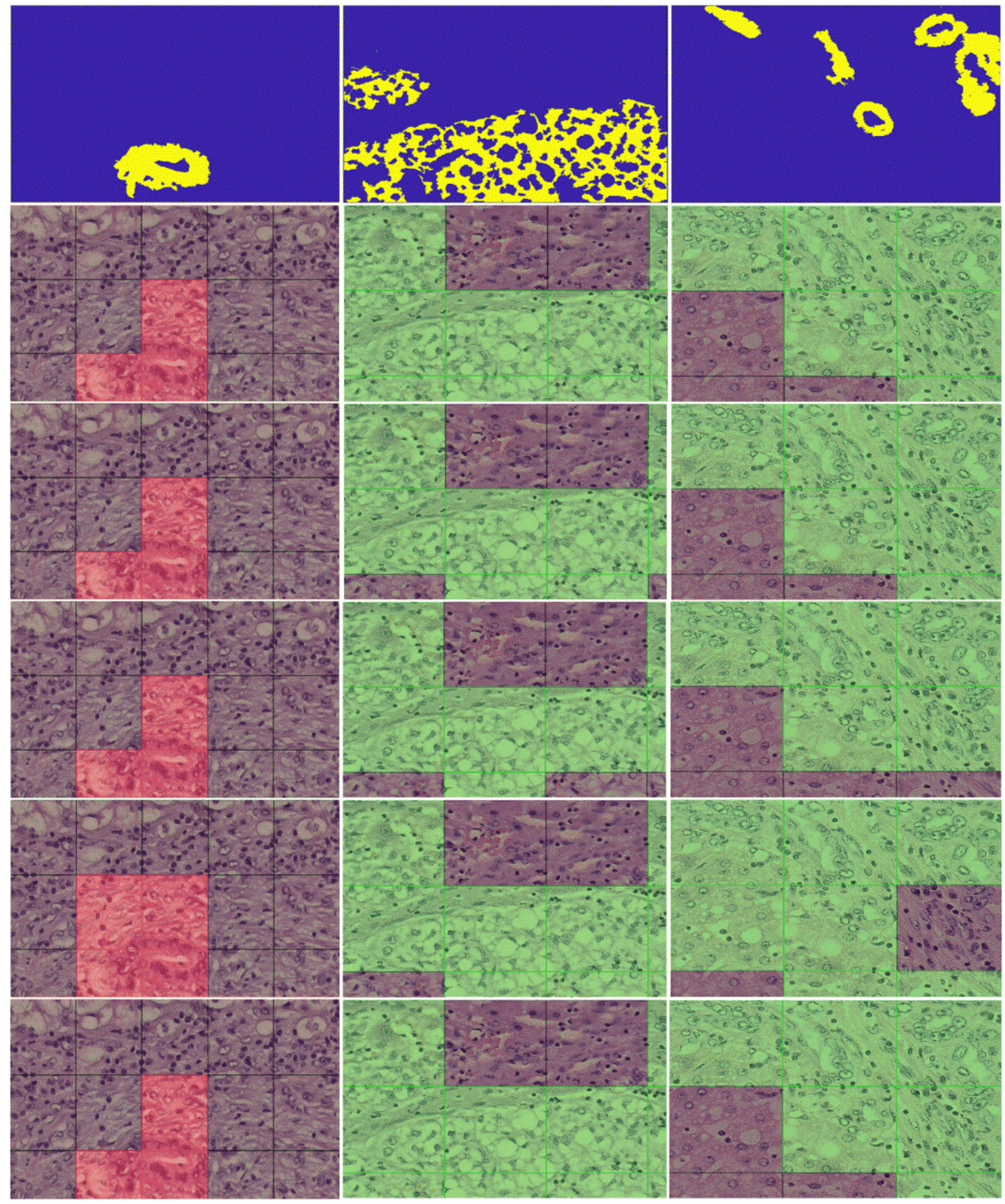

**Fig. 4**. Illustration of the diagnostic pipeline using deep networks trained on patch-level annotated images. Columns correspond to images diagnosed as colorectal metastatic adenocarcinoma (CMA), hepatocellular carcinoma (HCC), and cholangiocellular carcinoma (CCA), respectively. The first row shows ground-truth annotations obtained by majority voting from pixel-level annotations provided by four pathologists (cancer regions in yellow and background in blue) with at least two out of four pathologists had to agree on cancer label. Rows two to six present patch-level diagnostic maps generated by ConvNeXtV2, DenseNet, EfficientNetV2, MaxViT, and SwinV2, respectively. Color-coded patches indicate predicted cancer classes (red—CMA, green—HCC, blue—CCA) and are overlaid on the original H&E-stained images. Evidently, all networks incorrectly diagnosed CCA as HCC.

## 4. Results and discussion

We conducted experiments to evaluate whether nnU-Net–based semantic segmentation provides competitive performance for image-level diagnosis of liver cancers compared with deep networks trained on patch-level annotated images, as well as with diagnostic performance reported for immunohistochemical markers in the literature. Diagnostic performance was evaluated at multiple levels: (i) pixel level using nnU-Net semantic segmentation outputs; (ii) patch level obtained by aggregating nnU-Net pixel-level predictions; and (iii) image level obtained by aggregating nnU-Net semantic segmentation maps. In addition, image-level diagnosis was evaluated by aggregating patch-level predictions produced by five deep neural networks trained on patch-level annotated images: ConvNeXtV2, DenseNet, EfficientNetV2, MaxViT, and SwinV2. The experiments were performed using pixel-wise annotated images of hepatocellular carcinoma (HCC) (55 images from 30 patients), cholangiocellular carcinoma (CCA) (55 images from 29 patients), and colorectal metastatic adenocarcinoma (CMA) (60 images from 30 patients). Furthermore, we also evaluated the trained nnU-Net model on an additional external set of clinically diagnosed but non-annotated images consisting of 22 HCC images from 11 patients, 21 CCA image from 11 patients, and 59 CMA images from 39 patients. Finally, we present qualitative semantic segmentation results obtained by the nnU-Net model on several challenging histopathological images that were difficult to delineate even for experienced pathologists.

### *4.1. Training and test protocols*

All networks were trained using a five-fold cross-validation scheme. With the exception of one fold in the CCA dataset, which contained images from five patients, all other folds included images from six patients. In the HCC dataset, one fold contained nine images, one contained ten images, and three contained twelve images. In the CCA dataset, one fold contained eleven images, one contained eight images, and three contained twelve images. For the CMA dataset, each of the five folds contained twelve images. Overall, the dataset was reasonably well balanced across folds. No overlap between folds was allowed at either the image level or the patient level, ensuring that test data were not seen by the inference models during training. For the entire dataset, we employed nested cross-validation, resulting in five outer folds and ten inner folds. For each outer fold, an ensemble model was constructed from the ten models trained on the corresponding inner folds. For the external clinical dataset, five ensemble models (one from each outer fold) were used, and the final diagnostic label was obtained using a majority voting scheme.

### 4.2. Software environment

All deep learning–based classifiers were implemented using the PyTorch software framework (Paszke et al., 2019) on A100 GPU with 80 GB of local memory. All five patch-level classifiers are instantiated via the `timm` library (Wightman, 2019). Stain normalization and ground-truth evaluation were implemented in MATLAB (version 2024b) on a computer running a 64-bit Microsoft Windows 10 operating system. The system was equipped with 256 GB of RAM and an Intel Xeon E5-2650 v4 processor operating at a clock speed of 2.2 GHz. It took approximately 1150 GPU hours, i.e., 48 GPU days, to train six networks elaborated before in a nested cross-validation scheme.

### 4.3. Performance metrics

In our experiments, we used six metrics to evaluate the performance of the diagnostic models. First, we computed confusion matrices for each model, from which we derived the numbers of true positives (TP), true negatives (TN), false positives (FP), and false negatives (FN). Based on these values, we calculated sensitivity (SE), also known as recall or the true positive rate; specificity (SP), also referred to as the true negative rate; accuracy (ACC); balanced accuracy (BACC); precision, also known as the positive predictive value (PPV); and the F1 score (or Dice coefficient). All metrics range from 0 to 1, where 0 indicates the worst performance and 1 indicates the best performance. Accuracy estimates the probability that the model prediction is correct. Balanced accuracy represents the average probability that an instance of each class is correctly classified. Sensitivity measures the ability of the model to correctly identify positive instances in the dataset. Positive predicted value (a.k.a. precision) indicates the reliability of the model when predicting a positive instance. Specificity measures the ability of the model to correctly identify negative instances. Finally, the F1 score represents the harmonic mean of positive predicted value and sensitivity, providing a balanced measure of both metrics.

### 4.4. Results on pixel-level and patch-level annotated images

Table 2 presents the aggregate diagnostic results for all three liver cancer types obtained by the nnU-Net model and five deep neural networks on the described dataset. The nnU-Net produces predictions at the pixel level. For this level, performance metrics were computed from confusion matrices by comparing predicted labels with the ground-truth annotations, thereby estimating pixel-level diagnostic performance. For patch-level analysis, we assigned the dominant pixel label to patches of size 384×384 pixels. The same procedure was applied to the ground-truth segmentation maps. Performance metrics were then computed from confusion matrices at the patch level. A patch was assigned the background (BCKG) label only if all pixels within the patch were labeled as non-cancerous. For image-level diagnosis, the dominant label within the entire image was assigned as the final prediction, effectively treating the region of interest (ROI) as a single large patch. Performance metrics were computed from confusion matrices using the image-level ground-truth labels (clinical diagnoses). Since the five deep networks were trained using patch-level annotations, their predictions were also generated at the patch level. Image-level diagnoses were therefore obtained by assigning the dominant predicted patch label to the entire image. Performance metrics were then computed from confusion matrices with respect to the image-level ground-truth labels. As shown in Table 2, aggregation of predictions from the pixel level to the patch level, and subsequently to the image level, leads to improved performance metrics. This observation supports our hypothesis that aggregating semantic segmentation outputs into image-level diagnoses improves diagnostic robustness and reduces sensitivity to local annotation inconsistencies and domain variability. In contrast, the diagnostic performance achieved by the deep networks trained on patch-level labels further highlights the limitations of the patch-based labeling approach.

**Table 2.**

Comparison of diagnostic performance between the nnU-Net semantic segmentation-based approach and patch-level classifiers (ConvNeXtV2, DenseNet, EfficientNetV2, MaxViT, and SwinV2) trained on patch-annotated images. The best results are indicated in **bold**, the second-best results are <u>underlined</u>, and the third-best results are shown in *italic*.

| Algorithm | Diagnosis level | ACC | BACC | SE | SP | F1 | PPV |
|---|---|---|---|---|---|---|---|
| nnU-Net | image | **0.9667** | **0.9750** | **0.9667** | **0.9833** | **0.9666** | **0.9667** |
| nnU-Net | patch | <u>0.9467</u> | <u>0.9649</u> | <u>0.9477</u> | <u>0.9822</u> | *0.9439* | <u>0.9467</u> |
| nnU-Net | pixel | *0.9218* | *0.9218* | *0.8830* | *0.9604* | <u>0.9876</u> | *0.9218* |
| ConvNeXtV2 | image | 0.8667 | 0.8631 | 0.8645 | 0.9561 | 0.8624 | 0.8627 |
| DenseNet | image | 0.8765 | 0.9157 | 0.8728 | 0.9586 | 0.8752 | 0.8821 |
| EfficientNetV2 | image | 0.8672 | 0.9078 | 0.8501 | 0.9520 | 0.8495 | 0.8496 |
| MaxVit | image | 0.8550 | 0.9078 | 0.8501 | 0.9520 | 0.8495 | 0.8496 |
| SwinV2 | image | 0.8735 | 0.9134 | 0.8694 | 0.9579 | 0.8702 | 0.8729 |

Table 3 presents diagnostic results for the CMA class. The nnU-Net achieves perfect performance at the image level across all evaluation metrics. For this model, diagnostic performance improves progressively when predictions are aggregated from the pixel level to the patch level and finally to the image level. This observation again supports our hypothesis that aggregating semantic segmentation outputs into image-level diagnoses improves diagnostic robustness and reduces sensitivity to local annotation inconsistencies and domain variability. Deep networks generate predictions at the patch level, and image-level diagnoses are obtained by assigning the dominant predicted patch label to the entire image. Depending on the evaluation metric, their performance exceeds that of the nnU-Net when the latter is evaluated at the patch level after aggregation from pixel-level predictions. This observation further supports the notion that even aggregation from patch-level predictions to image-level diagnosis improves overall diagnostic performance.

**Table 3.**

Comparison of diagnostic performance of the CMA class between the nnU-Net semantic segmentation-based approach and patch-level classifiers (ConvNeXtV2, DenseNet, EfficientNetV2, MaxViT, and SwinV2) trained on patch-annotated images. The best results are indicated in **bold**, the second-best results are underlined, and the third-best results are shown in *italic*.

| Algorithm | Diagnosis level | ACC | BACC | SE | SP | F1 | PPV |
|---|---|---|---|---|---|---|---|
| nnU-Net | image | **1.0000** | **1.0000** | **1.0000** | **1.0000** | **1.0000** | **1.0000** |
| nnU-Net | patch | 0.9475 | 0.9820 | 0.9837 | 0.9824 | 0.9285 | 0.9603 |
| nnU-Net | pixel | 0.8138 | 0.9437 | 0.9085 | 0.9789 | 0.8973 | 0.8864 |
| ConvNeXtV2 | image | 0.9948 | *0.9561* | 0.9143 | 0.9979 | 0.9529 | 0.9948 |
| DenseNet | image | 0.9812 | 0.9517 | 0.9111 | 0.9922 | 0.9449 | 0.9812 |
| EfficientNetV2 | image | 0.9746 | 0.9511 | 0.9127 | 0.9895 | 0.9426 | 0.9746 |
| MaxVit | image | 0.9781 | 0.9558 | *0.9286* | 0.9909 | 0.9485 | 0.9781 |
| SwinV2 | image | *0.9897* | 0.9550 | 0.9143 | *0.9958* | *0.9505* | *0.9897* |

Table 4 presents diagnostic results for the HCC class. The nnU-Net achieves convincingly best performance at the image level across almost all evaluation metrics. It is also highly competitive with the performance at the patch level. This observation again supports our hypothesis that aggregating semantic segmentation outputs into the patch-level and then image-level diagnoses improves diagnostic robustness and reduces sensitivity to local annotation inconsistencies and domain variability. Deep networks generate predictions at the patch level, and image-level diagnoses are obtained by assigning the dominant predicted patch label to the entire image. Their performance is mostly inferior to the performance achieved by semantic segmentation nnU-Net model.

**Table 4.**

Comparison of diagnostic performance of the HCC class between the nnU-Net semantic segmentation-based approach and patch-level classifiers (ConvNeXtV2, DenseNet, EfficientNetV2, MaxViT, and SwinV2) trained on patch-annotated images. The best results are indicated in **bold**, the second-best results are underlined, and the third-best results are shown in *italic*.

| Algorithm | Diagnosis level | ACC | BACC | SE | SP | F1 | PPV |
|---|---|---|---|---|---|---|---|
| nnU-Net | image | **0.9091** | **0.9750** | **1.0000** | 0.9500 | **0.9594** | 0.9091 |
| nnU-Net | patch | *0.8968* | 0.9843 | 0.9924 | **0.9762** | 0.9456 | *0.9030* |
| nnU-Net | pixel | 0.7386 | *0.9321* | *0.8881* | *0.9761* | 0.8497 | 0.8144 |
| ConvNeXtV2 | image | 0.8645 | 0.8887 | 0.8220 | 0.9555 | 0.8427 | 0.8645 |
| DenseNet | image | 0.8992 | 0.8977 | 0.8277 | 0.9678 | *0.8619* | **0.9331** |
| EfficientNetV2 | image | 0.8791 | 0.8869 | 0.8125 | 0.9614 | 0.8445 | 0.8791 |
| MaxVit | image | 0.8220 | 0.8865 | 0.8352 | 0.9378 | 0.8289 | 0.8228 |
| SwinV2 | image | 0.8623 | 0.9035 | 0.8452 | 0.9528 | 0.8582 | 0.8623 |

Table 5 presents diagnostic results for the CCA class. The nnU-Net achieves convincingly best performance at the image level across all but one evaluation metrics. It is also very competitive with the performance at the patch level. This observation again supports our hypothesis that aggregating semantic segmentation outputs into the patch-level and then image-level diagnoses improves diagnostic robustness and reduces sensitivity to local annotation inconsistencies and domain variability. Deep networks generate predictions at the patch level, and image-level diagnoses are obtained by assigning the dominant predicted patch label to the entire image. Their performance is basically the third best where the first-best and second-best performances are achieved by semantic segmentation nnU-Net model.

**Table 5.**

Comparison of diagnostic performance of the CCA class between the nnU-Net semantic segmentation-based approach and patch-level classifiers (ConvNeXtV2, DenseNet, EfficientNetV2, MaxViT, and SwinV2) trained on patch-annotated images. The best results are indicated in **bold**, the second-best results are underlined, and the third-best results are shown in *italic*.

| Algorithm | Diagnosis level | ACC | BACC | SE | SP | F1 | PPV |
|---|---|---|---|---|---|---|---|
| nnU-Net | image | **0.9000** | **0.9500** | 0.9000 | **1.0000** | **0.9474** | **0.9091** |
| nnU-Net | patch | 0.8663 | 0.9522 | **0.9161** | *0.9833* | 0.9283 | 0.9400 |
| nnU-Net | pixel | 0.6977 | 0.8952 | 0.7973 | 0.9931 | 0.8219 | 0.8481 |
| ConvNeXtV2 | image | 0.8138 | 0.9129 | *0.8669* | 0.9589 | 0.8395 | 0.8138 |
| DenseNet | image | *0.8916* | *0.9172* | 0.8555 | 0.9788 | *0.8732* | *0.8916* |
| EfficientNetV2 | image | 0.8941 | 0.8965 | 0.8130 | 0.9800 | 0.8516 | 0.8841 |
| MaxVit | image | 0.8162 | 0.8956 | 0.8300 | 0.9612 | 0.8230 | 0.8162 |
| SwinV2 | image | 0.8617 | 0.9094 | 0.8470 | 0.9718 | 0.8543 | 0.8617 |

Table 6 presents diagnostic results for the background (BCKG) class. Since there were no images that were free of cancerous pixels, we evaluated prediction performance for nnU-Net on the pixel-level and patch-level, and prediction performance of other deep networks on patch-level only. To label patch as BCKG class all the pixels within the patch had to be labeled as non-cancerous. As it was the case with other classes, the nnU-Net achieves convincingly best performance across all evaluation metrics.

**Table 6.**

Comparison of diagnostic performance of the background (BCKG) class between the nnU-Net semantic segmentation-based approach and patch-level classifiers (ConvNeXtV2, DenseNet, EfficientNetV2, MaxViT, and SwinV2) trained on patch-annotated images. The best results are indicated in **bold**, the second-best results are <u>underlined</u>, and the third-best results are shown in *italic*.

| Algorithm | Diagnosis level | ACC | BACC | SE | SP | F1 | PPV |
|---|---|---|---|---|---|---|---|
| nnU-Net | patch | <u>0.8666</u> | **0.9405** | <u>0.8987</u> | **0.9824** | <u>0.9285</u> | **0.9603** |
| nnU-Net | pixel | **0.8964** | <u>0.9161</u> | **0.9381** | 0.8914 | **0.9454** | <u>0.9527</u> |
| ConvNeXtV2 | patch | 0.7776 | 0.8834 | 0.8548 | *0.9120* | 0.8144 | 0.7776 |
| DenseNet | patch | 0.7566 | 0.8963 | *0.8971* | 0.8956 | *0.8209* | 0.7566 |
| EfficientNetV2 | patch | 0.7485 | *0.8967* | 0.9026 | 0.8908 | 0.8183 | 0.7485 |
| MaxVit | patch | *0.7813* | 0.8661 | 0.8143 | <u>0.9179</u> | 0.7975 | *0.7813* |
| SwinV2 | patch | 0.7778 | 0.8667 | 0.8621 | 0.9113 | 0.8118 | 0.7778 |

### *4.5. Comparison with immunohistochemistry*

In Section 2.3, we presented a brief overview of the use of immunohistochemical (IHC) markers for differentiating hepatocellular carcinoma (HCC), cholangiocellular carcinoma (CCA), and metastatic adenocarcinoma (MA). To contextualize the results achieved by the nnU-Net and other deep learning models, we reproduce in Table 7 the sensitivity values reported in Table 1 of Chan and Yeh (2010), which summarizes nine IHC markers commonly used for differentiating HCC, CCA, and MA. These values are compared with the sensitivity and specificity results obtained by the nnU-Net and the patch-based deep networks. Ideally, an IHC marker that is highly sensitive for one cancer type should exhibit minimal sensitivity for the other types, which would correspond to high specificity. However, reported sensitivities of IHC markers vary considerably across studies. For example, Geramizadeh et al. (2007) reported that Hep Par 1 exhibited 86% sensitivity for HCC and 0% sensitivity for both CCA and MA. In contrast, Chu et al. (2002) reported 92% sensitivity for HCC, but also 7% sensitivity for CCA and 6% sensitivity for MA. Similarly, Yamauchi et al. (2005) reported 84% sensitivity of glypican-3 (GPC3) for HCC and 0% sensitivity for CCA, whereas Ligato et al. (2008) reported 83% sensitivity for HCC and 6% sensitivity for MA. Maeda et al. (1996) reported 97% sensitivity of CK7 for CCA, with sensitivities of 7% and 3% for HCC and MA, respectively. Other IHC markers reported in the literature exhibit varying levels of sensitivity across multiple liver cancer types. In comparison, the nnU-Net model with image-level diagnosis achieved sensitivities of 100% for HCC, 99% for CCA, and 100% for CMA, with corresponding specificities of 95%, 100%, and 100%, respectively. Deep networks trained on patch-level annotated images achieved sensitivity levels comparable to those reported for many IHC markers, although their overall diagnostic performance remained lower than that of the nnU-Net–based approach. Above results suggest that the nnU-Net method may assist pathologists in prioritizing immunohistochemical marker selection and could complement established diagnostic workflows.

**Table 7**

Sensitivities of various immunohistochemical (IHC) markers for detecting hepatocellular carcinoma (HCC), cholangiocellular carcinoma (CCA), and metastatic adenocarcinoma (MA). The sensitivity values are reproduced from Table 1 in Chan and Yeh (2010). Reference numbers correspond to the following studies: 1 – Geramizadeh et al. (2007), 2 – Lau et al. (2002), 3 – Chu et al. (2002), 4 – Fan et al. (2003), 5 – Morrison et al. (2002), 6 – Leong et al. (1998), 7 – Wang et al. (2006), 8 – Wang et al. (2008), 9 – Libbrecht et al. (2006), 10 – Shafizadeh et al. (2008), 11 – Ligato et al. (2008), 12 – Yamauchi et al. (2005), 13 – Coston et al. (2008), 14 – Wang et al. (2006), 15 – Man et al. (2005), 16 – Kandil et al. (2007), 17 – Stroescu et al. (2006), 18 – Ma et al. (1993), 19 – Proca et al. (2000), 20 – Maeda et al. (1996), 21 – Nascimento et al. (2007). *a* Positive pCEA staining is considered canalicular in HCC and membranous and/or cytoplasmic in CCA and MA. *b* Positive CD34 staining is defined as a complete or diffuse staining pattern of sinusoidal endothelial cells. Sensitivity (SE) and specificity (SP) for nnU-Net and five deep networks are also presented.

| Method | HCC | CCA | MA |
|---|---|---|---|
| Hep Par 1 | 86%[1], 90%[2], 92%[3], 95%[4], 96%[5] | 0%[2], 0%[5], 0%[1], 7%[3], 9%[4], 12.5%[6] | 0%[5], 2%[1], 6%[3], 14%[2] |
| GPC3 | 75%[7], 76%[8], 77%[9], 79%[10], 83%[11], 84%[12], 88%[13] | 0%[12], 10%[7], 19%[15] | 0%[16], 6%[11] |
| pCEA[a] | 50%[17], 69%[2], 71%[18], 96%[5] | 100%[2], 100%[5] | 93%[5], 96%[2] |
| MOC 31 | 0%[19], 7%[5], 12%[2], 14%[1] | 67%[2], 93%[5], 100%[19] | 66%[2], 85%[5], 92%[1], 100%[19] |
| CK7 | 7%[20], 20%[17], 21%[2] | 78%[2], 97%[20], 100%[17] | 3%[20], 36%[2] |
| CK8/18 | 70%[17] | 20%[17] | |
| CK19 | 0%[20], 10%[2] | 44%[2], 77%[20], 80%[17] | 29%[2], 64%[20] |
| CK20 | 0%[20], 5%[2] | 10%[20], 11%[2] | 30%[2], 74%[20] |
| CK34[b] | 94%[21], 95%[13] | | |
| nnU-Net (SE) | 100% | 99% | 100% |
| nnU-Net (SP) | 95% | 100% | 100% |
| ConvNeXtV2 (SE) | 82.20% | 86.69% | 91.43% |
| ConvNeXtV2 (SP) | 95.55% | 95.89% | 99.79% |
| DenseNet (SE) | 82.77% | 85.55% | 91.11% |
| DenseNet (SP) | 96.78% | 97.78% | 99.22% |
| EfficientNetV2 (SE) | 81.25% | 81.30% | 91.27% |
| EfficientNetV2 (SP) | 96.14% | 98.00% | 98.95% |
| MaxVit (SE) | 83.52% | 83.00% | 92.86% |
| MaxVit (SP) | 93.78% | 96.12% | 99.09% |
| SwinV2 (SE) | 84.52% | 84.70% | 91.43% |
| SwinV2 (SP) | 95.28% | 97.18% | 99.58% |

*4.6. Results on external clinical test images*

To further assess the diagnostic performance of the trained nnU-Net model, an independent test set was collected, comprising 59 CMA images, 22 HCC images from 11 patients, and 21 CCA image from 11 patients. Among the CMA images, 40 (from 20 patients) were acquired from formalin-fixed paraffin-embedded (FFPE) slides, while 19 (from 19 patients) were obtained intraoperatively using rapid tissue freezing, followed by cryotome sectioning and brief H&E staining (Sitnik et al., 2021). We compared the nnU-Net model, using aggregation from pixel-level predictions to the image level, with classifiers trained on patch-level annotated images (ConvNeXtV2, DenseNet, EfficientNetV2, MaxViT, and SwinV2). Diagnostic performance was evaluated using accuracy, F1 score, recall, and precision (Fig. 5). Both overall performance across all classes and class-specific metrics (CMA, HCC, and CCA) are reported. Additionally, CMA performance is stratified into two subcategories: CMA-FFPE and CMA-Intra. The nnU-Net model consistently outperforms the competing methods in terms of accuracy, F1 score, and precision across all classes. For recall, inferior performance is observed for CCA in comparison with diagnoses obtained by deep networks trained on patch-level annotated images. However, precision of nn-Unet in diagnosis of CCA class is significantly greater in comparison with diagnoses obtained by deep networks trained on patch-level annotated images.  It implies that nn-Unet diagnosis of CCA is more trustable and it is a consequence of small number of false positive CCA diagnoses generated by nn-Unet. The F1 score, that provides a balanced measure of both recall and precision, is in case of CCA class comparable for all networks. Notably, nnU-Net achieves strong performance for CMA, both overall and across subcategories. In contrast, competing methods show substantially reduced performance on CMA-Intra, which was not represented in the training set. Despite this domain shift, nnU-Net maintains high diagnostic performance, indicating improved robustness with respect to other competing methods. To further improve the presentation of diagnostic performance on the external clinical test set, we report confusion matrices in Tables 8 and 9 for the nnU-Net and SwinV2 models, respectively. As shown in Table 8, the nnU-Net model misclassified only 2 of 19 CMA-Intra samples as CCA. As opposed to that as shown in Table 9, the ConvNeXtV2 model misclassified 13 of 19 CMA-Intra samples as CCA, with one sample classified as background (BCKG) and one as HCC. Similar trends are observed for the other classifiers trained on patch-level annotated images. We conjecture that semantic segmentation and aggregation from pixel to image level mitigated the impact of domain shift.

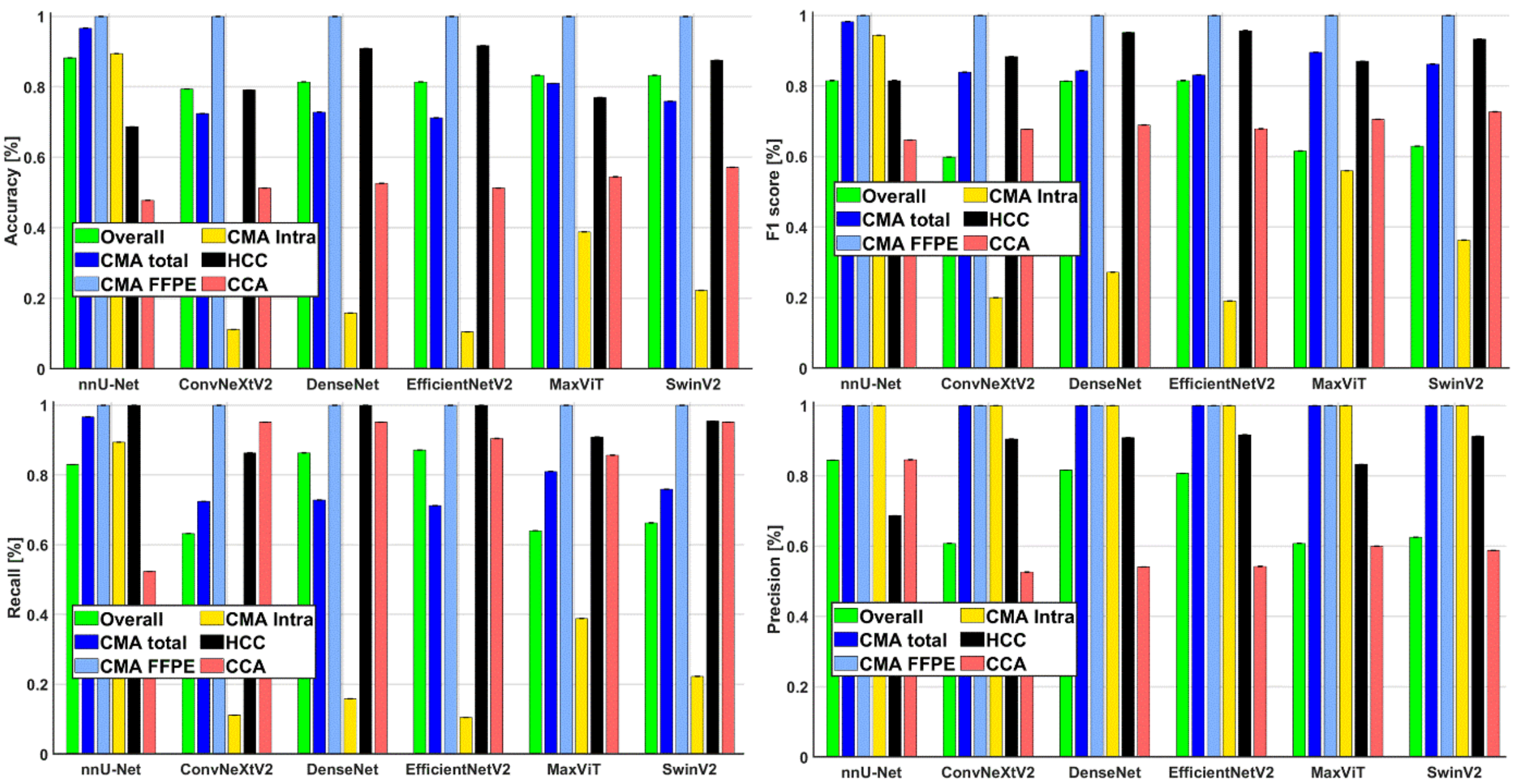

**Fig. 5**. Diagnostic performance of six deep learning models on an external clinical test set. Top row (left to right): accuracy and F1 score. Bottom row (left to right): recall and precision. The best performance equals to 1 and the worse to 0. Legend: Overall—all categories; CMA total—colorectal metastatic adenocarcinoma, including both FFPE and intraoperatively acquired sections; CMA-FFPE—CMA from FFPE sections only; CMA-Intra—CMA from intraoperatively collected sections only; HCC—hepatocellular carcinoma; CCA—cholangiocellular carcinoma.

**Table 8**

Confusion matrix generated by nnU-Net model on external clinical test set. Legend: BCKG - background (healthy) class; CMA-total - colorectal metastatic adenocarcinoma, including both FFPE and intraoperatively collected sections; CMA-FFPE—CMA from FFPE sections only; CMA-Intra—CMA from intraoperatively acquired sections only; HCC—hepatocellular carcinoma; CCA—cholangiocellular carcinoma.

| | BCKG | CMA | HCC | CCA |
|---|---|---|---|---|
| BCKG | 0 | 0 | 0 | 0 |
| CMA-total | 0 | 57 | 0 | 2 |
| CMA-FFPE | 0 | 40 | 0 | 0 |
| CMA-Intra | 0 | 17 | 0 | 2 |
| HCC | 0 | 0 | 22 | 0 |
| CCA | 0 | 0 | 10 | 11 |

**Table 9**

Confusion matrix generated by SwinV2 model on external clinical test set. Legend is the same as for Table 8.

| | BCKG | CMA | HCC | CCA |
|---|---|---|---|---|
| BCKG | 0 | 0 | 0 | 0 |
| CMA-total | 1 | 44 | 1 | 13 |
| CMA-FFPE | 0 | 40 | 0 | 0 |
| CMA-Intra | 1 | 4 | 1 | 13 |
| HCC | 0 | 0 | 21 | 1 |
| CCA | 0 | 0 | 1 | 20 |

### *4.7. Demonstration on several clinically challenging cases*

As described in Section 3.1, an initial set of 60 images per group was collected. Pixel-level annotations were obtained from four pathologists, and inter-annotator agreement was assessed for each image using Fleiss' kappa. Images with low agreement ($\kappa < 0.3$ for HCC and $\kappa < 0.4$ for CCA) were excluded from the main dataset. Fig. 6 presents two representative challenging cases and illustrates the diagnostic and localization performance of the trained nnU-Net model. In the HCC example, the model yields a correct diagnosis, and its segmentation closely matches the annotations of pathologists 3 and 4, while the regions annotated by pathologists 1 and 2 show no overlap with these areas. This example highlights that nnU-Net can support diagnostic decision-making by providing precise localization consistent with a subset of expert annotations. In the CCA example, the model produces an incorrect diagnosis; however, its segmentation shows strong agreement with the annotation of pathologist 2 and partial agreement with that of pathologist 4. This suggests that accurate localization, even in the presence of misclassification, may assist pathologists in identifying relevant regions and contribute to increased diagnostic confidence.

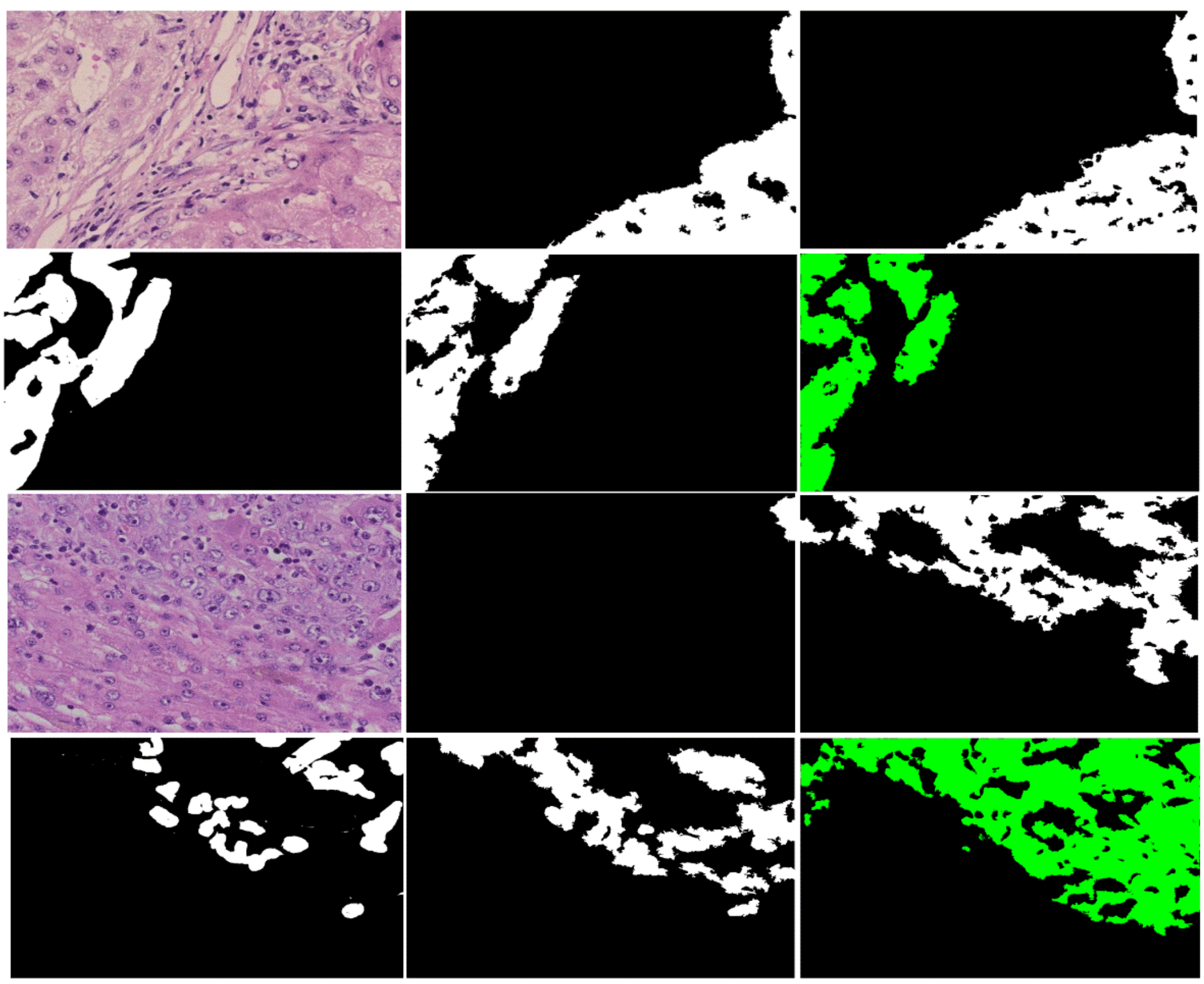

**Fig. 6**. Illustration of challenging cases for annotation, together with the localization and diagnostic performance of the trained nnU-Net model. The first two rows correspond to hepatocellular carcinoma (HCC), and the last two rows to cholangiocellular carcinoma (CCA). From left to right: H&E-stained image, pixel-wise annotations by four pathologists, and nnU-Net semantic segmentation with predicted diagnosis. In the annotation maps, white denotes tumor and black denotes background. Color-coded pixels in the model output indicate predicted classes (green: HCC; blue: CCA). Notably, the CCA case is misclassified by the model.

## 5. Conclusion

In this study, we introduced a publicly available dataset comprising 170 images of H&E-stained histopathological sections across three liver cancer categories: hepatocellular carcinoma (HCC; 55 images from 30 patients), cholangiocellular carcinoma (CCA; 55 images from 29 patients), and colorectal metastatic adenocarcinoma (CMA; 60 images from 30 patients). Pixel-level annotations were independently provided by four experienced pathologists, and consensus ground truth was established via majority voting with at least two out of four pathologists had to agree on cancer label. To the best of our knowledge, this is the first publicly available dataset offering pixel-wise annotations for these three liver cancer types. Additionally, we presented an external clinical test set, not used during training, consisting of 102 images of H&E-stained sections: HCC (22 images from 11 patients), CCA (21 image from 11 patients), and CMA (59 images, including 40 FFPE sections from 20 patients and 19 intraoperative samples from 19 patients).

We further proposed a semantic segmentation–driven framework for image-level diagnosis based on nnU-Net, an automated deep learning framework tailored for biomedical image segmentation. The approach derives image-level predictions by aggregating pixel-wise segmentation outputs. Under the clinically motivated assumption that each specimen corresponds to a single tumor type, the final diagnosis is assigned based on the most prevalent predicted class. Experimental results demonstrate that hierarchical aggregation—from pixel to patch to image level—consistently improves performance metrics. These findings support our hypothesis that aggregating semantic segmentation outputs enhances diagnostic robustness and reduces sensitivity to local annotation inconsistencies and domain variability. Moreover, the results align with known immunohistochemical patterns of differentiation among HCC, CCA, and CMA, suggesting potential clinical relevance.

Evaluation on the external clinical dataset further confirmed the robustness of the proposed approach. Notably, nnU-Net achieved strong performance for CMA across both FFPE and intraoperative fixed and frozen sections. In contrast, competing methods exhibited a substantial performance drop on intraoperative CMA samples, which were not represented in the training data. Despite this domain shift, nnU-Net maintained high diagnostic performance, indicating improved generalization capability relative to alternative approaches.

Finally, qualitative analysis on diagnostically challenging cases demonstrated that the model can produce segmentations closely matching annotations from at least one expert pathologist. These results highlight the potential of the proposed method to assist in identifying diagnostically relevant regions, thereby supporting clinical interpretability and increasing diagnostic confidence. Future work will focus on further evaluating the robustness of the proposed framework to variations in image acquisition conditions, including differences in microscopes and scanning devices.

**CrediT authorship contribution statement**

**Ivica Kopriva:** Writing - original draft, Conceptualization, Methodology, Supervision, Funding acquisition, MATLAB-based software for stain normalization. **Dario Sitnik:** Software for pixel-wise annotations, software for deep networks training on A100 GPU, Writing - original draft, review & editing. **Arijana Pačić:** Data curation, Pixel wise annotation, Writing - review & editing. **Irena Veliki Dalić:** Data curation, Pixel wise annotation, Writing -review & editing. **Ana Lukač:** Pixel wise annotation, Writing - review & editing. **Karolina Krstanac:** Pixel wise annotation, Writing-review & editing. **Marijana Popović Hadžija:** Data curation, Writing - review & editing.

**Declaration of competing interest**

The authors declare that they have no known competing financial interests or personal relationships that could have appeared to influence the work reported in this paper.

**Data and code availability**

Pythorch code with explanations is available at: `https://github.com/dsitnik/liver3c`. Weights of trained networks are available at: `https://data.fulir.irb.hr/data/HandE-Liver3C/HandE-Liver3C_weights.zip`. The target image together with stain normalization code in MATLAB script language is available at: `https://github.com/dsitnik/liver3c/tree/main/scripts/preprocessing/stain_normalization`.

Data are available at**:** `https://data.fulir.irb.hr/data/HandE-Liver3C/HandE-Liver3C_data.zip`.

**Acknowledgment**

This work was supported by the Croatian Science Foundation under project number HRZZ-IP-2022-10-6403.